\documentclass[10pt,twocolumn,letterpaper]{article}

\usepackage{cvpr}
\usepackage{times}
\usepackage{epsfig}
\usepackage{graphicx}
\usepackage{amsmath}
\usepackage{bm}
\usepackage{amssymb}
\usepackage{ltablex}
\usepackage{stmaryrd}
\usepackage{mathtools}
\usepackage[title]{appendix}

\newcommand{\matr}[1]{\bm{#1}}     % ISO complying version
\newcommand{\haofu}[1]{{{#1}}}

\newcommand{\RNum}[1]{\uppercase\expandafter{\romannumeral #1\relax}}

\newcommand{\best}[1]{{\bf{#1}}}

\DeclareMathOperator{\G}{G}
\DeclareMathOperator{\D}{D}

\DeclareMathOperator{\softmax}{softmax}

\DeclareMathOperator{\relu}{LReLU}
\DeclareMathOperator{\mlp}{mlp}
\DeclareMathOperator{\gap}{gap}

\usepackage[pagebackref=true,breaklinks=true,letterpaper=true,colorlinks,bookmarks=false]{hyperref}

\cvprfinalcopy % *** Uncomment this line for the final submission

\def\cvprPaperID{4745} % *** Enter the CVPR Paper ID here

\ifcvprfinal\fi
\begin{document}

%%%%%%%%% TITLE
\title{Example-Guided Scene Image Synthesis using Masked Spatial-Channel Attention and Patch-Based Self-Supervision}

\author{Haitian Zheng
\quad Haofu Liao \quad Lele Chen \quad Wei Xiong \quad Tianlang Chen \quad Jiebo Luo\\
University of Rochester\\
{\tt\small \{hzheng15, hliao6, lchen63, wxiong5, tchen45, jluo\}@cs.rochester.edu}
}

\maketitle
%\thispagestyle{empty}

%%%%%%%%% ABSTRACT
\begin{abstract}
    Example-guided image synthesis has been recently attempted to synthesize an image from a semantic label map and an exemplary image.
    In the task, the additional exemplar image serves to provide style guidance that control the appearance of the synthesized output.
    Despite the controllability advantage, the previous models are designed on datasets with specific and roughly aligned objects. In this paper, we tackle a more challenging and general task, where the exemplar is an arbitrary scene image that is semantically unaligned to the given label map. To this end, we first propose a new Masked Spatial-Channel Attention (MSCA) module which models the correspondence between two unstructured scenes via cross-attention.
    Next, we propose an end-to-end network for joint global and local feature alignment and synthesis. In addition, we propose a novel patch-based self-supervision scheme to enable training. Experiments on the large-scale CCOO-stuff dataset show significant improvements over existing methods. Moreover, our approach provides interpretability and can be readily extended to other tasks including style and spatial interpolation or extrapolation, as well as other content manipulation.
\end{abstract}

\section{Introduction}
\label{sec:intro}
\begin{figure}[]
	\centering
	\includegraphics[width=0.88\linewidth]{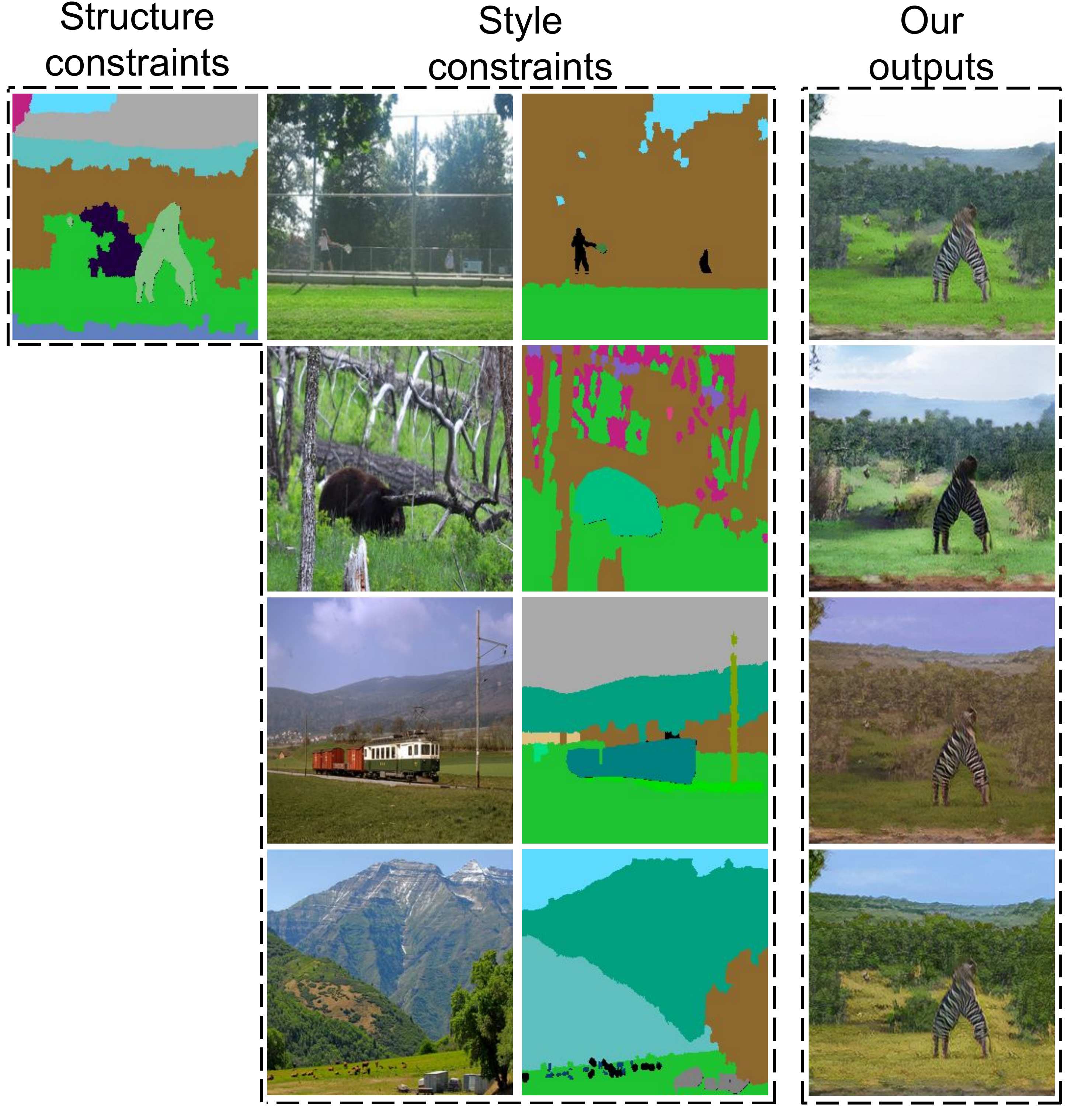}
	\caption{The inputs to style-consistent scene image generation is a  \emph{structurally uncorrelated} and \emph{semantically unaligned} segmentation map (column 1) and a reference image (column 2) that constraints the style of the output. The corresponding reference segmentation map is also taken as input. In spite of the complexity of the task, our model can generate high-quality scene images with a consistent style with the reference image.
	%\todo{change image, better result. add an arrow, so that it looks like 3:4}
    }
	\label{fig:teaser}
\end{figure}

Conditional generative adversarial network (cGAN)~\cite{conditionalGAN} has recently made substantial progresses in realistic image synthesis. In cGAN, a generator $\hat{x}=\G(c)$ aims to output a realistic image $\hat{x}$ with a constraint implicitly encoded by $c$. Conversely, a discriminator $\D(x,c)$ learns such a constraint from ground-truth pairs $\langle x,c\rangle$ by predicting if $\langle \hat{x},c\rangle$ is real or generated. 

The current cGAN models~\cite{spade,pix2pixhd,pix2pix} for semantic image synthesis aim to solve the \emph{structural consistency} constraint where 
% \haofu{$c$ is a semantic label map and the output $x$ is required to structurally aligned with $c$.}
% \haofu{The limitation of the direct semantic image synthesis is that the styles of the output images are inherently decided by the model and cannot controlled by users.} 
% still uncontrollable to users, as users are agnostic about the style of outputs.
the output image $x$ is required to be aligned to a semantic label map $c$.
The limitation of the above generative process is that the styles of the image outputs are inherently determined by the model and thus cannot be controlled by users. To provide desired controllability over the generated styles, previous studies~\cite{example_cvpr18,example_cvpr19} 
\haofu{impose additional constraints and allow more inputs to the generator:} $\hat{x}_{2 \rightarrow 1} =\G(z, c_1, x_2)$,
% \zht{allow for an extra exemplar image to guides the output styles. Formally,}
where $x_2$ is an exemplar image that guides the style of $c_1$. However, previous studies are designed on datasets such as face~\cite{liu2015deep,rossler2018faceforensics}, dancing~\cite{example_cvpr19} or street view~\cite{yu2018bdd100k}, where the input images usually contain similar semantics and the spatial structures of $y$ and $c_x$ are usually similar as well.

\haofu{Different from the previous studies, we propose to address a more challenging example-guided \emph{scene image} generation task. As shown in Fig.~\ref{fig:teaser}, given a semantic label map $c_1$ (column 1) and an arbitrary scene image $x_2$ (column 2) with its semantic map $c_2$ (column 3) as the input, the task aims to generate a new scene image $\hat{x}_{2 \rightarrow 1}$ (column 4) that matches the semantic structure of $c_1$ and the scene style of $x_2$. The challenge is that scene images have complex semantic structures as well as diversified scene styles, and more importantly, the inputs $c_1$ and $x_2$ are \emph{structurally uncorrelated} and \emph{semantically unaligned}. Therefore, a mechanism is required to better match the structures and semantics for coherent outputs, e.g., the tree styles can be applied to mountains but cannot be applied to sky.}

%\zht{Different from the previously studied tasks~\cite{example_cvpr18,example_cvpr19,example_iclr19}, we study a more challenging example-guided \emph{scene image} generation task. Our task takes an arbitrary scene image with its semantic label map (in Fig.~\ref{fig:teaser} column 2 and 3, respectively) as exemplar-guidance to synthesize a new scene image (in Fig.~\ref{fig:teaser} column 4). Ours inputs are both \emph{structurally uncorrelated} and \emph{semantically unaligned}. Hence, a better mechanism is required to propagate information.}

% the reference images (column 3) contains high-quality visual at each semantic regions. However, the reference images are structurally unrelated to the given conditions (column 1) and contains multiple new semantic regions while missing others. As a result, a more sophisticated mechanism than feature concatenation is demanded to propagate information from the reference images.

\haofu{In this paper, we propose a novel Masked Spatial-Channel Attention (MSCA) module (Section~\ref{subsect:spatialchannel}) to propagate features across unstructured scenes. Our module is inspired by a recent work~\cite{doubleattention} for attention-based object recognition, but we apply a new cross-attention approach to model the semantic correspondence for image synthesis instead. To facilitate example-guided synthesis, we further improve the module by including: i) feature masking for semantic outlier filtering, ii) multi-scaling for global and local feature processing, and iii) resolution extending for image synthesis. As a result, our module provides both clear physical meaning and interpretability for the example-guided synthesis task.}

%\zht{In this paper, we propose novel Masked Spatial-Channel Attention (MSCA) module (Section~\ref{subsect:spatialchannel}) to propagate features across unstructured scenes. Our module is inspired by a recent work~\cite{doubleattention} for attention modeling. Differently, we apply cross-attention to model correspondence for image synthesis. Moreover, improvements are made in the following aspects: i) feature masking for semantic outlier filtering, ii) multi-scaling for better feature aggregation, and iii) the extended resolution for image synthesis. Finally, our module owns clear physical meaning and interpretability for the example-guided synthesis task.} 

\haofu{We formulate the proposed approach under an unified synthesis network for joint feature extraction, alignment and image synthesis. We achieve this by applying MSCA modules to the extracted features for multi-scale feature domain alignment. Next, we apply a recent feature normalization technique, SPADE~\cite{spade} on the aligned features to allow spatially-controllable synthesis. To facilitate the learning of this network, we propose a novel patch-based self-supervision scheme. As opposed to~\cite{example_cvpr19}, our scheme requires only semantically parsed images for training and does not rely on video data.
% for semantic matched pairs $c_1$ and $x_2$. 
We show that a model trained with this approach generalizes over scales and across different scene semantics.
}

%\zht{Next, we propose an end-to-end synthesis network (Figure~\ref{fig:generator}) for joint feature extraction, alignment and image synthesis. We achieves this by applying MSCA modules on the extracted features for multi-scale feature domain alignment. We apply a recent feature normalization technique, SPADE~\cite{spade} on the aligned features to allow spatially-controllable synthesis.}

%\zht{Finally, we propose a novel patch-based self-supervision scheme to enable training. Acquiring style-consistent pairs is known hard. Recent works on style-guided image-to-image translation~\cite{munit,lee2018diverse,example_iclr19} apply disentangle learning which separates style and content. However, it is less studied how to disentangle style and content from both semantic label maps and images. A more recent work~\cite{example_cvpr19} utilize videos to generate paired data. However, acquiring diversified scene videos can be hard. Instead, our self-supervision scheme requires only annotated images for training. We show that model trained with our data scheme generalizes over scales and across different scene semantics.}

%\vspace{10pt}
Our main contributions include the following:
   \vspace{-0.08cm}
\begin{itemize}
    \item A novel masked spatial-channel attention (MSCA) module to propagate features for unstructured scenes. %Improvements over~\cite{doubleattention} are made including feature masking, multi-scaling and extended resolution.
    \vspace{-0.08cm}
    \item An unified synthesis network for joint feature extraction, alignment and image synthesis. %MSCA modules are applied for multi-scale feature domain alignment. A recent feature normalization techniques~\cite{spade} is used for spatially-controllable synthesis.
    \vspace{-0.08cm}
    \item A novel patch-based self-supervision scheme that requires only annotated images for training. 
    % \wei{what is the advantage of self-supervision training? I cannot get it from your introduction.}
    \vspace{-0.08cm}
    \item Experiments on COCO-stuff~\cite{cocostuff} dataset that show significant improvements over existing methods. Moreover, our model provides interpretability and can be extended to other tasks of content manipulation.
\end{itemize}

\section{Related work}
\noindent \textbf{Generative Adversarial Networks.} \quad
Recent years have witnessed the progresses of generative adversarial networks (GANs)~\cite{gan} for image synthesis. A GAN model consists of a generator and a discriminator where the generator serves to produce realistic images that cannot be distinguished from the real ones by the discriminator. Recent techniques for realistic image synthesis include 
modified losses~\cite{wasserstein,lsgan,improved},
model regularization~\cite{sn},
self-attention~\cite{sagan,largescalegan}, feature normalization~\cite{stylegan} and progressive synthesis~\cite{progressivegan}.

\noindent \textbf{Image-to-Image translation (I2I).} \quad
I2I translation aims to translate images from a source domain to a target domain. The initial work of Isola~\etal~\cite{pix2pix} proposes a conditional GAN framework to learn I2I translation with paired images. Wang~\etal~\cite{pix2pixhd} improve the conditional GAN for high-resolution synthesis and content manipulation. 
To enable I2I translation without using paired data, a few works~\cite{cyclegan,liu2017unsupervised,munit,drit,pairedcyclegan} apply the cycle consistency constraint in training. Recent works on photo-realistic image synthesis take semantic label maps as inputs for image synthesis. Specifically, Wang~\etal~\cite{pix2pixhd} extend the conditional GAN for high-resolution synthesis, Chen~\etal~\cite{CRN} propose a cascade refine pipeline. More recently, Park~\etal~\cite{spade} propose spatial-adaptive normalization for realistic scene image generation.

\noindent \textbf{Example-Guided Style Transfer and Synthesis.} \quad
Example guided style transfer~\cite{image_analogies,Image_quilting} aims to transfer the style of an example image to a target image. More recent works~\cite{gatys2016image,adaptive_instance_normalization,phototransfer,johnson2016perceptual,deep_image_analogy,feature_shuffle,pairedcyclegan,video_style_transfer,wct2} utilize deep neural network features to model and transfer styles. Several frameworks~\cite{munit,huang2018multimodal,example_iclr19} perform style transfer via image domain style and content disentanglement. In addition, domain adaptation~\cite{pairedcyclegan} applies a cycle consistency loss to cross-domain style transformation.

More recently, example-guided synthesis~\cite{example_cvpr18,example_cvpr19} is proposed to transfer the style of an example image to a target condition, e.g. a semantic label map. Specifically, Lin~\etal~\cite{example_cvpr18} apply dual learning to disentangle the style for guided synthesis, Wang~\etal~\cite{example_cvpr19} extract style-consistent data pairs from videos for model training. 
% In addition, Wang~\etal~\cite{pix2pixhd} and Park~\etal~\cite{spade} adopt I2I networks to self-encoding versions for example-guided style transfer.
In addition, Park~\etal~\cite{spade} adopt I2I networks to self-encoding versions for example-guided style transfer.
Different from~\cite{example_cvpr18,example_cvpr19,spade}, we address spatial alignment of complex scenes for better {\it style integration in multiple regions of an image}. Furthermore, our patch-based self-supervision learning scheme does not require video data and is a general version of self-encoding.

%  Lin~\etal~\cite{example_cvpr18} apply dual learning for $64\times64$-resolution synthesis. 
% However, a modern generator like SPADE~\cite{spade} takes long times to train. Therefore, scaling dual learning can be challenging due to the doubled training time and potential gradient vanishing problems or destabilized training. Recently, Wang~\etal~\cite{example_cvpr19} instead generate paired and unpaired frames from video data. However, scene images are are less structured thus, inherently more challenging than the face, dancing or street view images used in~\cite{example_cvpr19}. Furthermore, current scene video datasets are less diverse or scaled than the available scene datasets~\cite{cocostuff}. In this paper, we propose a novel patch-based supervision to tackle the deficiency of paired scene images. We show that the patch-based supervision is generalized, and it can transfer to global scene images across vastly different semanti\noindent

\noindent \textbf{Correspondence Matching for Synthesis.} \quad
Finding correspondence is critical for many synthesis tasks. For instance, Siarohin~\etal~\cite{Deformable} apply the affine transformation on reference person images to improve pose-guided person image synthesis, Wang~\etal~\cite{vid2vid} use optical flow to align frames for coherent video synthesis. However, the affine transformation and optical flow cannot adequately model the correspondences between two arbitrary scenes.

The recent self-attention~\cite{wang2018non,sagan} can capture general pair-wise correspondences. However, self-attention is computationally intensive at high-resolution. Later, Chen~\etal~\cite{doubleattention} propose to factorize self-attention for efficient video classification. Inspired by \cite{doubleattention}, we propose an attention-based module named MSCA. It is worth noting MSCA is based on cross-attention and feature masking for spatial alignment and image synthesis.
\section{Method}
The proposed approach aims to generate scene images that align with given semantic maps. Differ from conventional semantic image synthesis methods~\cite{pix2pix,pix2pixhd,spade}, our model takes an exemplary scene as an extra input to provide more controllability over the generated scene image. Unlike existing exemplar-base approaches~\cite{example_cvpr18,example_cvpr19}, our model addresses the more challenging case where the exemplary inputs are structurally and semantically unaligned with the given semantic map.

%\zht{Our model generates a scene image that is aligned to a semantic label map while matches the style of a reference scene. Differ from image-to-image translation task~\cite{pix2pix,pix2pixhd,spade}, our task takes an exemplary scene as extra inputs. And unlike~\cite{example_cvpr18,example_cvpr19}, the exemplary inputs can be arbitrary both structurally and semantically.}

\begin{figure}[]
	\centering
	\includegraphics[width=0.88\linewidth]{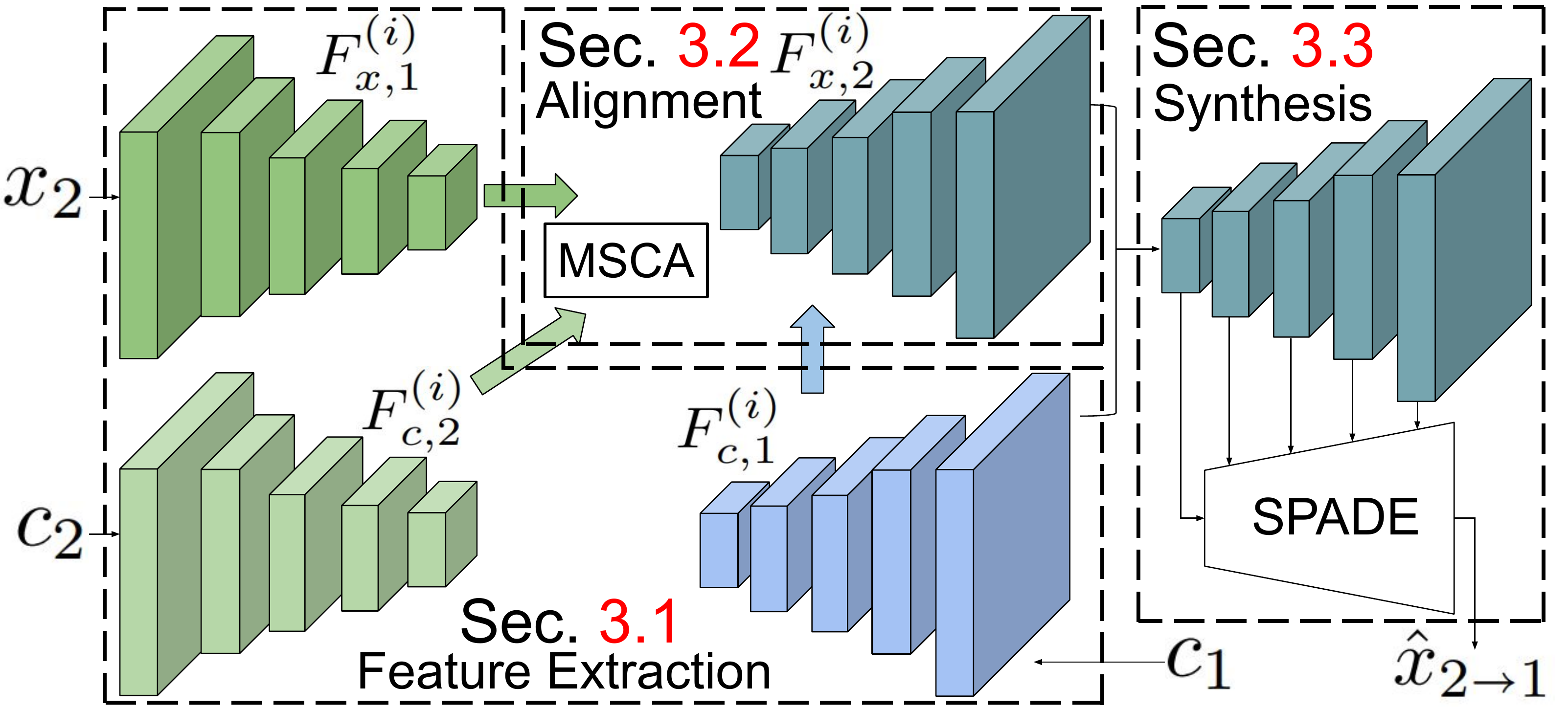}
	\caption{Our generator consists of three steps, namely i) feature extraction, ii) spatial feature alignment, and iii) image synthesis. We elaborate each step in its corresponding section, respectively.
	%\todo{change figure notation, add MSCA inside alignment, add line, change spade color(mix). change arrow}
	}
	\label{fig:generator}
\end{figure}

Our method takes a semantic map $c_1$, a reference image $x_2$ and its corresponding semantic map $c_2$ as inputs and synthesizes an image $\hat{x}_{2\shortrightarrow 1}$ which matches the style of $x_2$ and structure of $x_1$ using a generator $\G$, $\hat{x}_{2\shortrightarrow 1}=\G(c_1,x_2,c_2)$.
As shown in Fig.~\ref{fig:generator}, the generator $\G$ consists of three parts, namely i) feature extraction ii) feature alignment and iii) image synthesis. In Sec.~\ref{subsect:encode}, we describe the first part that extracts features from inputs of both scenes. In Sec.~\ref{subsect:spatialchannel}, we propose a masked spatial-channel attention (MSCA) module to distill features and discovery relations between two arbitrarily structured scene. Unlike the affine-transformation~\cite{stn} and flow-base warping~\cite{vid2vid}, MSCA provides a better interpretability to the scene alignment task. In Sec.~\ref{subsect:synthesis}, we introduce how to use the aligned features for image synthesis. Finally, in Sec.~\ref{subsect:patchsupervision}, we propose a patch-based self-supervision scheme to facilitate learning.

\subsection{Feature Extraction}
\label{subsect:encode}

Taking an image $x_2$ and label maps $c_1,c_2$ as inputs, the feature extraction module extracts multi-scale feature maps for each input. Specifically, the feature map $F^{(i)}_{x,2}$ of image $x_2$ at scale $i$ is computed by:
\begin{align}
\label{eq:image_feature}
\begin{aligned}
    F^{(i)}_{x,2} = W^{(i)}_x \ast F_{\text{vgg}}^{(i)}(x_2), \quad \text{for $i\in \{0, \dots, L\}$}, 
\end{aligned}
\end{align}
where $\ast$ denotes the convolution operation, $F_{\text{vgg}}^{(i)}$ denotes the feature map extracted by VGG-19~\cite{vgg} at scale $i$, and $W^{(i)}_x$ denotes a $1\times 1$ convolutional kernel for feature compression. $L$ is the number scales and we set $L=4$ in this paper.

For label map $c_1$, its feature $F^{(i)}_{c,1}$ is computed by:
\begin{align}
\label{eq:segment_feature}
    F^{(i)}_{c,1}=
    \begin{cases}
        \relu(W^{(i)}_{c} \ast c^{(i)}_1) & \text{for $i=L$}, \\
        \relu(W^{(i)}_{c} \ast [\Uparrow(F^{(i+1)}_{c,1}), c_1^{(i)}]) &\text{otherwise},
    \end{cases}
\end{align}
where $\Uparrow(\cdot)$ denotes $\times 2$ bilinear interpolation, $c^{(i)}_1$ denotes the resized label map, $W^{(i)}_{c}$ denotes a $1\times 1$ convolutional kernel for feature extraction, and operation $[\cdot, \cdot]$ denotes channel-wise concatenation. Note that as scale $i$ decreases from $L$ down to $0$, the feature resolutions in Eq.~\ref{eq:segment_feature} are progressively increased to match a finer label maps $c^{(i)}_1$.

Similarly, applying Eq.~\ref{eq:segment_feature} with the same weights to label map $c_2$, we can extract its features $F^{(i)}_{c,2}$:
\begin{align}
\label{eq:segment_feature2}
    F^{(i)}_{c,2}=
    \begin{cases}
        \relu(W^{(i)}_{c} \ast c^{(i)}_2) & \text{for $i=L$} \\
        \relu(W^{(i)}_{c} \ast [\Uparrow(F^{(i+1)}_{c,2}), c_2^{(i)}]) &\text{otherwise}
    \end{cases}.
\end{align}

\subsection{Masked Spatial-channel Attention Module}
\label{subsect:spatialchannel}
\begin{figure}[]
	\centering
	\includegraphics[width=0.88\linewidth]{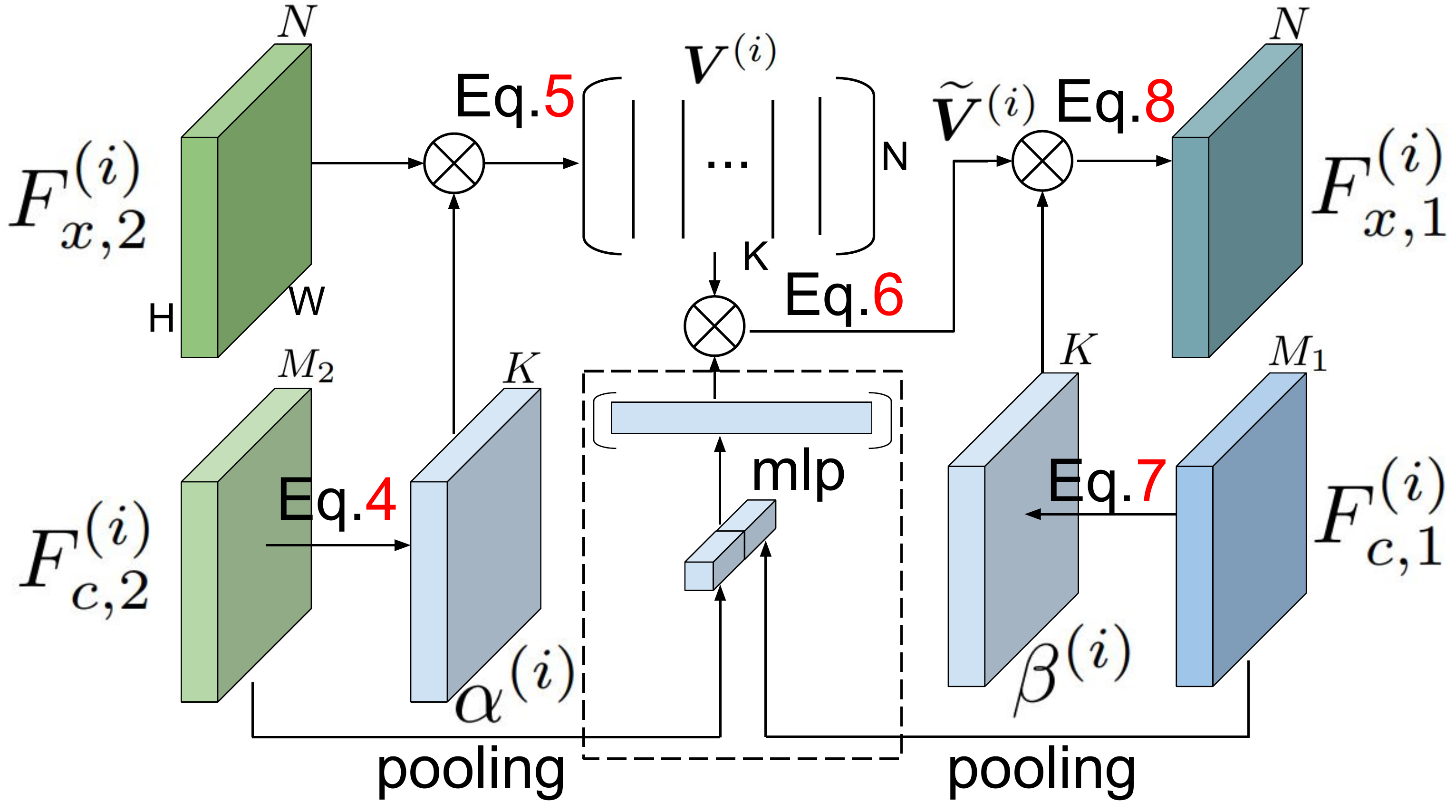}
	\caption{The spatial-channel attention module for feature alignment. Our module takes image features map $F^{(i)}_{x,2}$ and segmentation features map $F^{(i)}_{c,1}$, $F^{(i)}_{c,2}$ as inputs to output a new image feature map $F^{(i)}_{x,1}$ that is aligned to condition $c_1$.
	% \todo{change figure notation, (image, segmap, all features's name, ), element-wise multiplication, add spatial channel attention map to figure to illstrate. }
	}
	\label{fig:sc-attention}
\end{figure}

As shown in Fig.~\ref{fig:sc-attention}, taking the image features $F^{(i)}_{x,2}$ and the label map features $F^{(i)}_{c,1}$, $F^{(i)}_{c,2}$ as inputs\footnote{We assume spatial resolution at scale $i$ being $H\times W$ and channel size of $F^{(i)}_{x,2}$, $F^{(i)}_{c,1}$, $F^{(i)}_{c,2}$ being $N,M_1,M_2$, respectively. }, the MSCA module generates a new image feature map $F^{(i)}_{x,1}$ that has the content of $F^{(i)}_{x,2}$ but is aligned with $F^{(i)}_{c,1}$. We elaborate the detailed procedures as follows:

\noindent \textbf{Spatial Attention.} \quad
Given feature maps $F^{(i)}_{x,2}, F^{(i)}_{c,2}$ of the exemplar scene, the module first computes a spatial attention tensor $\alpha^{(i)}\in {[0,1]}^{K\cdot H\cdot W}$:
\begin{align}
\label{eq:spatial-attention}
\begin{aligned}
    \alpha^{(i)} = \softmax_{2,3}(\phi^{(i)} \ast [F^{(i)}_{x,2}, F^{(i)}_{c,2}]),
\end{aligned}
\end{align}
with $\phi^{(i)}\in\mathbb{R}^{(N+M_2) \cdot K}$ denoting a $1\times1$ convolutional filter and $\softmax_{2,3}$ denoting a 2D softmax function on spatial dimensions $\{2,3\}$. The output tensor contains $K$ attention maps of resolution $H\times W$, which serve to attend $K$ different spatial regions on image feature $F^{(i)}_{x,2}$.

\noindent \textbf{Spatial Aggregation.} \quad
Then, the module aggregates $K$ feature vectors from $F^{(i)}_{x,2}$ using the $K$ spatial attention maps of $\alpha^{(i)}$ from Eq.~\ref{eq:spatial-attention}. Specifically, a matrix dot product is performed:
\begin{align}
\label{eq:spatial-aggregate}
\begin{aligned}
    \matr{V}^{(i)} &= \matr{F}^{(i)}_{x,2} (\matr{\alpha}^{(i)})^T,
\end{aligned}
\end{align}
with $\matr{\alpha}^{(i)}\in [0,1]^{K\cdot HW}$ and $\matr{F}^{(i)}_{x,2}\in \mathbb{R}^{C\cdot HW}$ denoting the reshaped  versions of $\alpha^{(i)}$ and $F^{(i)}_{x,2}$, respectively. The output $\matr{V}^{(i)} \in \mathbb{R}^{C \cdot K} $ stores feature vectors spatially aggregated from the $K$ independent regions of $F^{(i)}_{x,2}$.

\noindent \textbf{Feature Masking.} \quad
The exemplar scene $x_2$ may contain irrelevant semantics to the label map $c_1$, and conversely, $c_1$ may contain  semantics that are unrelated to $x_2$. 
% However, with the above formulation, each output location must correspond to at least one region in exemplar scene.
To address this issue, we apply feature masking on the output of Eq.~\ref{eq:spatial-aggregate} by multiplying $\matr{V}^{(i)}$ with a length-$K$ gating vector at each row:
\begin{align}
\label{eq:masking}
\begin{aligned}
    \widetilde{\matr{V}}^{(i)} &= (\matr{V}^{(i)})^T \circ \mlp([\gap(F^{(i)}_{c,1}),\gap(F^{(i)}_{c,2})]),
\end{aligned}
\end{align}
where $\mlp(\cdot)$ denotes a 2-layer MLP followed by a sigmoid function, $\gap$ denotes a global average pooling layer, $\circ$ denotes broadcast element-wise multiplication, and $\widetilde{\matr{V}}^{(i)}$ denotes the masked features. The design of feature masking in Eq.~\ref{eq:masking} resembles to Squeeze-and-Excitation~\cite{SEnet}. Using the integration of global information from label maps $c_1$ and $c_2$, features are filtered.

\noindent \textbf{Channel Attention.} \quad
Given feature $F^{(i)}_{c,1}$ of label map $c_1$, a channel attention tensor $\beta^{(i)}\in {[0,1]}^{K\cdot H\cdot W}$ is generated as follows:
\begin{align}
\label{eq:channel-attention}
\begin{aligned}
    \beta^{(i)} = \softmax_{1}(\psi^{(i)} \ast F^{(i)}_{c,1}),
\end{aligned}
\end{align}
with $\psi^{(i)}\in\mathbb{R}^{M_1\cdot K}$ denoting a $1\times1$ convolutional filter and $\softmax_{1}$ denoting a softmax function on channel dimension. The output $\beta^{(i)}$ serves to dynamically reuse features from $\widetilde{\matr{V}}^{(i)}$.

\noindent \textbf{Channel Aggregation.} \quad
With channel attention $\beta^{(i)}$ computed in Eq.~\ref{eq:channel-attention}, feature vectors at $HW$ spatial locations are aggregated again from $ \widetilde{\matr{V}}^{(i)}$ via matrix dot product:
\begin{align}
\label{eq:channel-aggregate}
\begin{aligned}
    \matr{F}^{(i)}_{x,1} &=  \widetilde{\matr{V}}^{(i)} (\matr{\beta}^{(i)})^T,
\end{aligned}
\end{align}
where $\matr{\beta}^{(i)}\in \mathbb{R}^{K\cdot HW}$ denotes the reshaped version of $\beta^{(i)}$. The output $\matr{F}^{(i)}_{x,1} \in \mathbb{R} ^{N \cdot HW}$ represents the aggregated features at $HW$ locations. The output feature map $F^{(i)}_{x,1}$ is generated by reshaping $\matr{F}^{(i)}_{x,1}$ to size $ N \times H \times W$.

\noindent \textbf{Remarks.} \quad
Spatial attention (Eq.~\ref{eq:spatial-attention}) and aggregation (Eq.~\ref{eq:spatial-aggregate}) attend to $K$ independent regions from feature $F^{(i)}_{x,2}$, then store the $K$ features into $\matr{V}^{(i)}$.
After feature masking, given a new label map $c_1$, channel attention (Eq.~\ref{eq:spatial-attention}) and aggregation (Eq.~\ref{eq:channel-aggregate}) combine $\widetilde{\matr{V}}^{(i)}$ at each location to compute a output feature map. As results, each output location finds its correspondent regional features or ignored via feature masking. In this way, the feature of example scene is aligned. Note that when $K=1$ and $\alpha^{(i)}$ is constant, the above operations is essentially a global average pooling. We show in experiment that $K=8$ is sufficient to dynamically capture visually significant scene regions for alignment.

\begin{figure}[t]
	\centering
	\includegraphics[width=0.88\linewidth]{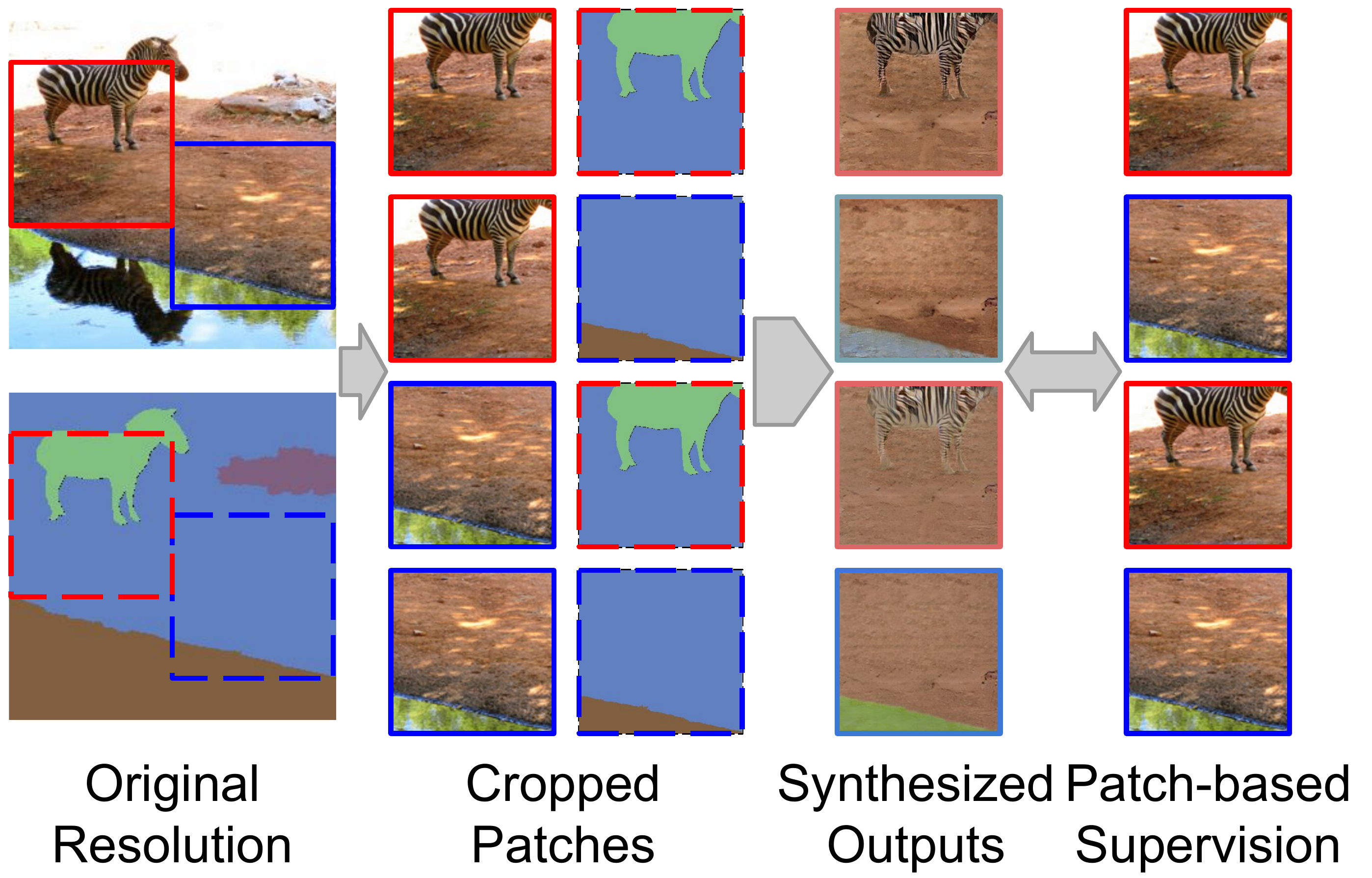}
	\caption{Our patch-based self-supervision scheme performs self-reconstruction and cross-reconstruction for two-patches from the same image.
	%\todo{Add notation}
	}
	\label{fig:patchsupervision}
\end{figure}

\noindent \textbf{Multi-scaling.} \quad
Both global color tone and local appearances are informative for the style-constraint synthesis. Therefore, we apply MSCA modules at all scales $i\in\{0,\dots,L\}$ to generate global and local features $F^{(i)}_{x,1}$. 
% In such a way, global and local appearance styles are extracted from the example scene image.

\begin{figure*}[t]
	\centering
	\includegraphics[width=1.0\linewidth]{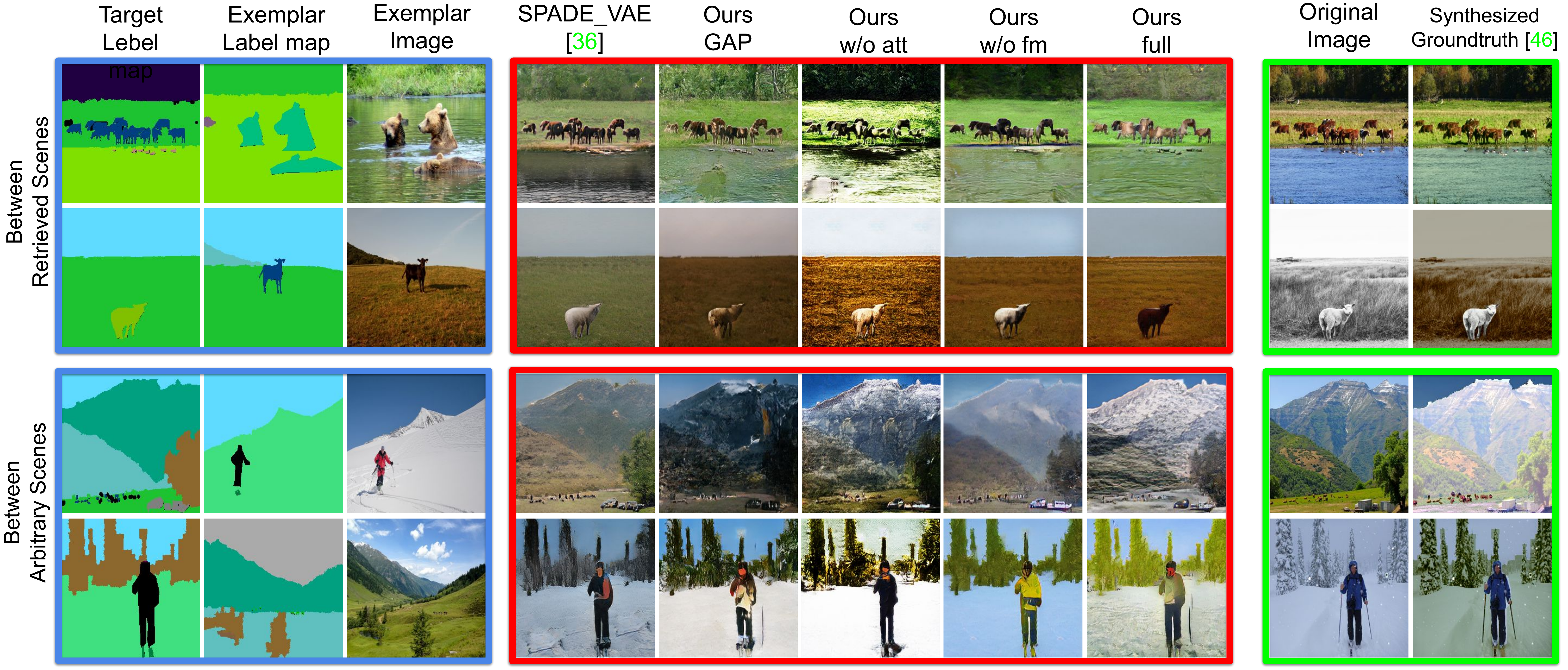}
	\caption{Visual comparisons with SPADE\_VAE, and ours ablation models. Example-guided scene synthesis is performed between two {\bf{retrieved scenes}} (rows 1,2) and two {\bf{arbitrary scenes}} (rows 3,4).  Columns 1 to 3 (blue) depict depict the target label maps, exemplar label maps and the associated images, respectively. Columns 4 to 8 in (red) depict different methods and our model (Columns 8).  Columns 9 and 10 (green) respectively depict original images from target label map and synthesized ground truth using~\cite{wct2} in the \emph{retrieved} dataset (top of Table~\ref{tab:main_experiment}). In comparison, our method clearly produces the most style-consistent ({\bf with the exemplar!}) and visually plausible images.
% 	\todo{change blue box to other color, bolder box.}
	}
	\label{fig:compare_main}
\end{figure*}

\subsection{Image Synthesis}
\label{subsect:synthesis}
The extracted features $F^{(i)}_{c,1}$ in Sec.~\ref{subsect:encode} capture the semantic structure of $c_1$, whereas the aligned features $F^{(i)}_{x,1}$ in Sec.~\ref{subsect:spatialchannel} capture the appearance style of the example scene. In this section, we leverage $F^{(i)}_{c,1}$ and $F^{(i)}_{x,1}$ as control signals to generate output images with desired structures and styles. 

Specifically, we adopt a recent synthesis model, SPADE~\cite{spade}, and feed the concatenation of $F^{(i)}_{x,1}$ and $F^{(i)}_{c,1}$ to the spatially-adaptive denormalization layer of SPADE at each scale. By taking the style and structure signal as inputs, spatially-controllable image synthesis is achieved. We refer readers to appendix for more network details of the synthesis module.

\subsection{Patch-Based Self-Supervision}
\label{subsect:patchsupervision}
Training a synthesis model requires style-consistent scene pairs. However, paired scenes are hard to acquire. To overcome the issue, we propose a patch-based self-supervision scheme which enables training.
% Previous work~\cite{example_cvpr19} relies on video data to generate supervision. However, there exist no outdoor scene video datasets at present. Instead, our generator is trained directly from image dataset. 

Our basic assumption is that if patches $x_p$ and $x_q$ come from the same scene, they share the same style. Consequently, using patch $x_p$ as exemplar, both $x_p$ and the other patch $x_q$ can be reconstructed, i.e. self-reconstruction and cross-reconstruction. More formally, we sample non-overlapping patches $\langle x_p,c_p \rangle$ and $\langle x_q,c_q \rangle$ at locations $p$ and $q$ from a same scene $\langle x,c \rangle$. To enable training, four images are synthesized in one training step:
\begin{align}
\label{eq:4image}
\begin{aligned}
    \hat{x}_{p\shortrightarrow p}&=\G(c_p,x_p,c_p),\\
    \hat{x}_{p\shortrightarrow q}&=\G(c_q,x_p,c_p),\\
    \hat{x}_{q\shortrightarrow p}&=\G(c_p,x_q,c_q),\\
    \hat{x}_{q\shortrightarrow q}&=\G(c_p,x_p,c_p),
\end{aligned}
\end{align}
and compared against groundtruths $x_p, x_q, x_p, x_q$. An illustrative example is shown in Fig.~\ref{fig:patchsupervision}. Note that patches $x_p, x_q$ do not necessary share the same semantics and our model is required to complete example-missing regions with reasonable content through learning. Our training objective is adopted from to~\cite{spade}. However, we apply pixel domain $\ell_1$ loss to encourage color consistency.
In our implementation, the generation processes in Eq.~\ref{eq:4image} share the same feature extraction, spatial attention, channel attention computation to reduce memory footprint during training.

\section{Experiments}
\label{sec:experiment}
\noindent \textbf{Dataset} \quad
Our model is trained on the \emph{COCO-stuff} dataset~\cite{cocostuff}. It contains densely annotated images captured from various scenes. We remove indoor images and images of random objects from the training/validation set, resulting in $21,648$/499 scene images for training/testing. 

During training, we resize images to $512 \times 512$ then crop two non-overlapping $256 \times 256$ patches to facilitate patch-based self-supervision. The two patches are cropped either in the left and right halves of the $512 \times 512$ image, or alternatively in the top and bottom halves.

The COCO-stuff dataset does not provide ground-truth for example-guided scene synthesis, i.e. two scene images with the exact same styles. To qualitatively evaluate model performances, we require a model to transfer the style from $x_2$ to $x_1$, where $x_2$ is the test image and $x_1$ is the generated image, in three ways: i) \emph{duplicating}: we use the test image itself to test self-reconstruction, ii) \emph{mirroring}: $x_1$ is generated by horizontally mirroring $x_2$, iii) \emph{retrieving}: $x_1$ is generated by finding the best match from the larger image pool. Specifically, we generate 20 candidate images from the training set with the smallest label histogram intersections. Out of the 20 images, the best-matching image $x_1$ is generated using SIFT Flow~\cite{siftflow}. Finally, since the color of $x_1$ and $x_2$ are not the same, we apply~\cite{wct2} on image $x_1$ for color correction. Examples of the \emph{retrieving} pairs are shown in Fig.~\ref{fig:compare_main}, in columns 3 and 10.

% we construct new dataset called \emph{COCO-stuff-match}. Specifically, we match each validation image with a training image using the following procedure. First, we remove the indoor scenes from the \emph{COCO-stuff} training/validation set, resulting 21648/499 scene images. Then, for each image from the validation set, we retrieve 20 candidate images from the training set using $\ell_1$ distance on segmentation histogram. Afterwards, a best-matched image out of the 20 image is generated using SIFT Flow~\cite{siftflow}. Examples of the and best-matched image is shown in Fig.~\ref{fig:dataset} \TODO{add figure}. Finally, since the style of the generated pairs are not identically same, we further perform photo-realistic style transfer~\cite{wct2} to generate the synthesis ground-truth, resulting 499 image pairs for training. More details of the dataset construction are included in appendix.

% Additionally, we also proposed the \emph{COCO-stuff-flip} dataset.  The \emph{COCO-stuff-flip} dataset uses an validation image as target, and the its flipped image as reference. The dataset measures how well model can reconstruct the groundtruth using its variant, i.e., with flipping.

\begin{table}[]
	\centering
	\resizebox{\columnwidth}{!}{
	\begin{tabular}{ |l|c|c|c|c|}
    %   \toprule
    %   \toprule
		\hline
		\hline
		Methods                             &	PSNR$\uparrow$ & SSIM$\uparrow$ &  LPIPS$\downarrow$	& FID$\downarrow$\\
		\hline
		{\em retrieving}   & & & &\\
		\hline
% 		Pix2pixHD\_IE~\cite{pix2pixhd}   & & & &\\
		SPADE\_VAE~\cite{spade}     &   15.62&  0.39& 0.480 &89.77\\
		ours GAP                    &   15.77&  0.39& 0.456 &89.55\\
		ours MSCA w/o att            & 11.76&  0.27  & 0.524& 98.35\\
		ours MSCA w/o fm	            & 15.64&  \best{0.40}  & 0.455& 89.58\\
		our full                & \best{15.98}&  \best{0.40}  & \best{0.449}& \best{85.87}\\
		\hline
		{\em mirroring}   & & & &\\
		\hline
% 		pix2pixHD\_IE~\cite{pix2pixhd}   & & & &\\
		SPADE\_VAE~\cite{spade}     &   15.72&  0.39& 0.478 &  89.58\\
		ours GAP                    &   16.06&  0.39& 0.446 & 89.54\\
		ours MSCA w/o att            & 12.13&  0.28  & 0.512& 98.02\\
		ours MSCA w/o fm	            & 16.52&  \best{0.42}  & 0.442& 88.40\\
		our full                & \best{16.95}&  \best{0.42}  & \best{0.425}& \best{83.20}\\
		\hline
		{\em duplicating}   & & & &\\
		\hline
% 		pix2pixHD\_IE~\cite{pix2pixhd}   & & & &\\
		SPADE\_VAE~\cite{spade}     &  15.35&  0.38& 0.476 &  90.69\\
		ours GAP                    &  15.70&  0.38& 0.438 & 88.51\\
		ours MSCA w/o att            &11.92&  0.28 & 0.508& 102.24\\
		ours MSCA w/o fm	            &15.91&  \best{0.40} & 0.437&89.44\\
		our full                &\best{16.50}&  \best{0.40} & \best{0.420}&\best{84.93}\\
		\hline
		\hline
% 	\bottomrule
	\end{tabular}
	}
	\caption{Quantitative comparisons of different methods in terms of PSNR, SSIM, LPIPS~\cite{lpips} and Fréchet Inception Distance (FID)~\cite{fid}. Higher scores are better for metrics with uparrow ($\uparrow$), and vice versa. 
% 	\todo{add pix2pixhd}
	}
	\label{tab:main_experiment}
\end{table}

\begin{figure*}[t]
	\centering
	\includegraphics[width=0.95\linewidth]{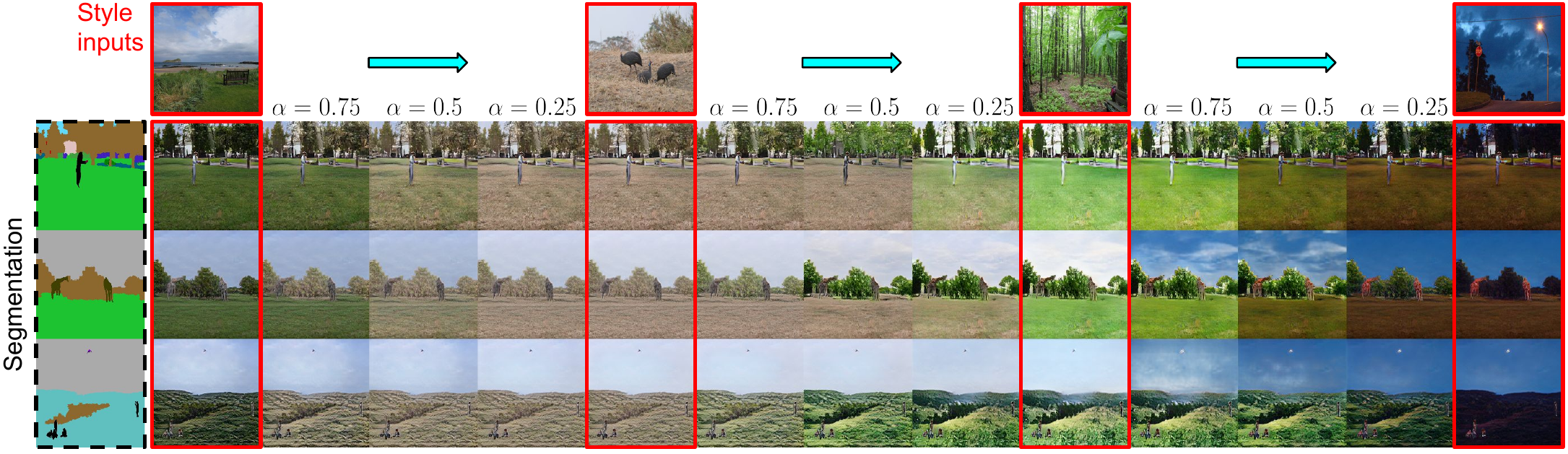}
	\caption{Style interpolation and traverse at test stage. We perform style traverse along path $\texttt{grass} \shortrightarrow \texttt{dessert}\shortrightarrow \texttt{forest}\shortrightarrow \texttt{night}$. Please refer to the Interpolation part in Sec.~\ref{sec:experiment} for details.}
	\label{fig:interpolate}
\end{figure*}

\noindent \textbf{Implementation Details} \quad
The number of attention maps $K$ for MSCA modules are set to $8,16,16,16,16$ from scale $0$ to $4$. The learning rate is set to $0.0002$ for the generator and the discriminator. The weights of generator are updated every $5$ iterations. We adopt the Adam~\cite{adam} optimizer ($\beta_1=0.9$ and $\beta_2=0.999$) in all experiments. Our synthesis model and all comparative models are trained for $20$ epochs to generate the results in the experiments. 
% For the ablation studies, all our models are trained for $20$ epochs. In comparison,  SPADE-VAE~\cite{spade} and Pix2pix-IE~\cite{pix2pixhd} (will be discussed later) are trained for $20$ epochs.

During implementation, we pretrain the spatial-channel attention with a lightweight feature decoder to avoid the ineffective but extremely slow updating of SPADE parameters. Specifically, at each scale, the concatenation of $F^{(i)}_{x,1}$ and $F^{(i)}_{c,1}$ in Sec.~\ref{subsect:synthesis} at each scale is fed into a $1\times 1$ convolutional layer to reconstruct the ground-truth VGG feature at the corresponding scale. The pretraining takes around ~$4$\% of the total training time to converge. More details of the pretraining procedure is provided in the appendix.

\noindent \textbf{Quantitative Evaluation} \quad
% We compare our approach with two realistic style-consistent synthesis approaches: instance-wise encoding Pix2pixHD (pix2pixHD\_IE)~\cite{pix2pixhd} and variational autoencoding SPADE (SPADE\_VAE)~\cite{spade}. Both pix2pixHD\_IE and SPADE\_VAE are based on self-reconstruction loss training. Therefore, we directly feed the resized $256\times 256$ images to train the two model. 
We compare our approach with an example-guided synthesis approach: variational autoencoding SPADE (SPADE\_VAE)~\cite{spade} which is based on a  self-reconstruction loss for training. Therefore, we directly use the resized $256\times 256$ images to train the model. We also attempt to train two example-guided synthesis models~\cite{example_cvpr18} and~\cite{example_cvpr19} (\cite{example_cvpr19} is trained using patch-based self-supervision) but cannot achieve visually good results. We leave the result of~\cite{example_cvpr18,example_cvpr19} in the  appendix. In addition, three ablation models are evaluated (see Ablation Study).

For quantitative evaluation, we apply low-level metrics including PSNR and SSIM~\cite{ssim}, and perceptual-level metrics including Perceptual Image Patch Similarity Distance (LPIPS)~\cite{lpips} and Fréchet Inception Distance (FID)~\cite{fid} on different models. For LPIPS, we use the linearly calibrated VGG model (see~\cite{lpips} for details).

As shown in Table \ref{tab:main_experiment}, our method clearly outperforms the remaining methods. Improvements in low-level and perceptual-level measurements suggest that our model better preserves color and texture appearances.
We observe that the performances of various approaches on the  \emph{retrieving} dataset are worse and less differentiated than their counterparts on the \emph{mirroring} and \emph{duplicating} datasets. It suggests that the \emph{retrieving} dataset is harder and noisier, as one cannot retrieve images that have the exact same styles. On \emph{retrieving} dataset, our approach achieves a moderate +0.36 PSNR gain over SPADE\_VAE (from 15.62 to 15.98).  By contrast, our approach achieves visually superior results over SPADE\_VAE on  \emph{duplicating} and \emph{mirroring}, e.g. +1.15 PSNR gain (from 15.35 to 16.50) on \emph{duplicating} and +1.23 PSNR gain (from 15.72 to 16.95) in PSNR on \emph{mirroring}.
%  \todo{the above numbers are mingled}
%%%%%%%%%%%%%
%\begin{figure}[]
%\centering
%\includegraphics[width=1.0\linewidth]{figures_tip/flowfashion-user-study.pdf}
%\caption{Subjective quality assessment of different algorithms. For each algorithm, the bar depicts the number of occurrences of scores, while blue to yellow colors represent the scores from the best to the worst.}
%\label{fig:user_study}
%% \vspace{-3mm}
%\end{figure}

\begin{figure*}[t]
	\centering
	\includegraphics[width=0.95\linewidth]{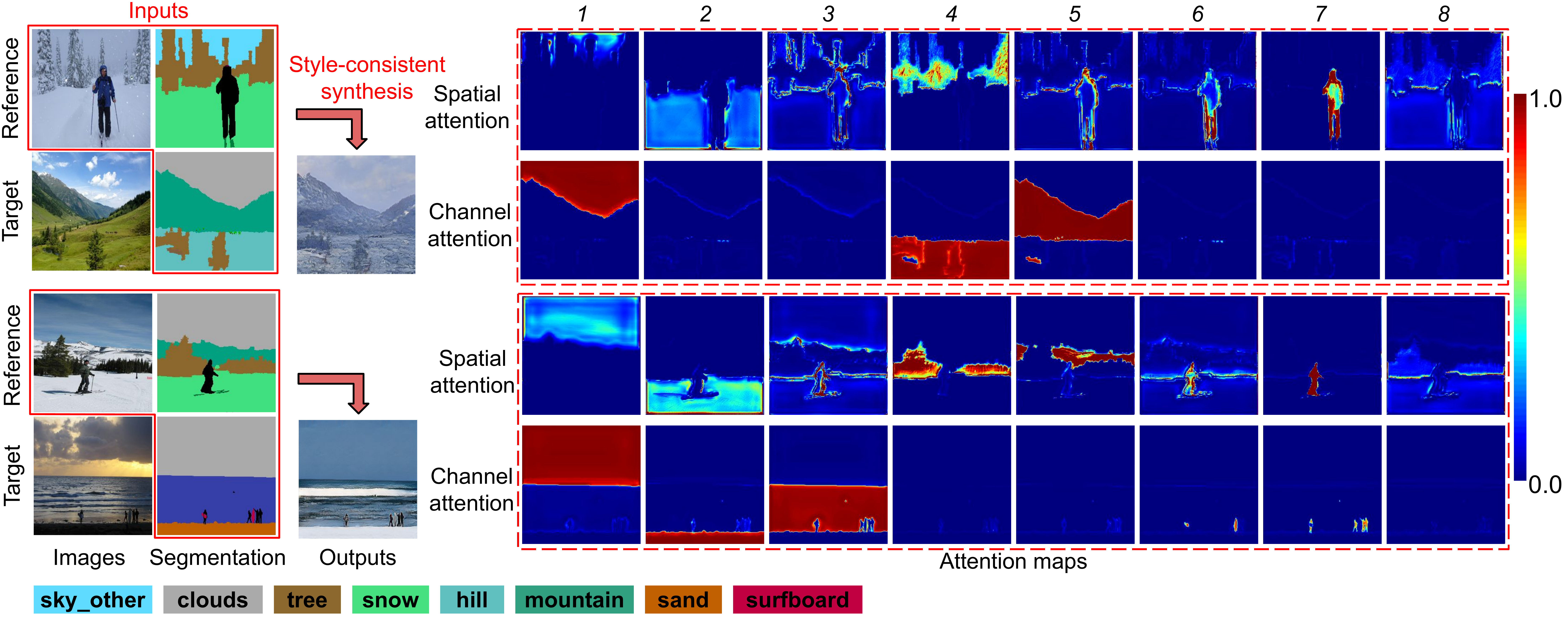}
	\caption{Right: the learned spatial attention and channel attention. Left: inputs and outputs of our model. Each spatial and channel attention attends to a specific region in the reference image and segmentation, respectively. By examining the segmentation semantics, we observe the following transformation patterns: $\texttt{sky\_other} \shortrightarrow \texttt{clouds}$, $\texttt{tree} \shortrightarrow \texttt{\{tree, hill\}}$ for row $1$, and $\texttt{clouds} \shortrightarrow \texttt{clouds}$, $\texttt{snow} \shortrightarrow \texttt{sand}$, $\texttt{other} \shortrightarrow \texttt{\{surfboard,other\}}$ for row $2$.
	%\todo{change heatmap legend}
	}
	\label{fig:attention}
\end{figure*}
\vspace{0.2cm}

\noindent \textbf{Qualitative Evaluation} \quad
Fig.~\ref{fig:compare_main} qualitatively compares our model against the remaining models on two {\bf{}{retrieved}} scenes (rows 1-2) and two {\bf{arbitrary}} scenes (rows 3-4). Our model achieves better style-consistent example-guided synthesis. Remarkably, in rows 3-4, even though the two scenes have very different semantics (indicated by the different colors of the corresponding label maps), our model can still maintain the styles of the exemplars while maintaining the correct semantics of the target label maps, e.g. generating ``snow'' rather than ``grass'' in row 4.

Also notice that sometimes our results are more style-consistent than the synthesized ground truths (last column). This further shows that the existing style transfer approach~\cite{styletransfer,wct2,phototransfer} cannot be directly applied to exemplar-guided scene synthesis for satisfactory results. 

% \noindent \textbf{Human Evaluation} \quad
% \TODO{Do the Evaluation}
% We conduct a subjective assessment to evaluate our method qualitatively. Specifically, we ask subjects to rank images among the 3 algorithms (\cite{pix2pixhd,spade} and ours) based on the style similarly and image quality. Table~\ref{} shows the evaluation result. 

% From the figure, our method is most frequently chosen as the best due to structurally consistent texture. DSCF \cite{deform_gan} achieves the second place due to its ability to maintain texture structure from the source image using rigid transformations. The qualitative results of different approaches as well as the warped source image and foreground/mask prediction from stage-\RNum{2} are shown in Fig.~\ref{fig:compare_main}. It can be noticed that the existed approaches generate blurry results or incorrect textures. By contrast, our method can preserve texture details from source images. Notably, our approach generates better warping results in comparison with IF, especially under large pose changes.

\noindent \textbf{Ablation Study} \quad
To evaluate the effectiveness of our design, we separately train three variants of our model: i) \textit{our GAP} that replaces the MSCA module with global average pooling,
ii) \textit{our MSCA w/o att} that keeps MSCA moduels but replaces spatial and channel attention of MSCA by one-hot label maps from source and target domains, respectively. In such way, alignment is performed on regions with the same semantic labeling, and iii) \textit{our MSCA w/o fm} that keeps MSCA modules but removes the feature masking procedures.
In Table~\ref{tab:main_experiment} and Fig.~\ref{fig:compare_main}, our model clearly achieves the best quantitative and qualitative results. In comparison, in Fig.~\ref{fig:compare_main}, \textit{our GAP} produces similar appearances in each region, as GAP cannot distinguish local appearances. \textit{Our w/o att} is less stable in training and cannot generate plausible results. We hypothesize that the label-level alignment will generate more misaligned and noisier feature maps, thus hurts training. \textit{our MSCA w/o fm} cannot perform correct appearance transformation, for instance,  transferring ``sky'' to ``snow'' (Fig.~\ref{fig:compare_main}, last row).

\noindent \textbf{The Effect of Attention} \quad
To understand the effect of spatial-channel attention, we visualize the learned spatial and channel attention in Fig.~\ref{fig:attention}. We observe that: 
a) spatial attention can attend to multiple regions of the reference image. For each reference region, channel attention finds the corresponding target region.
b) spatial-channel attention can detect and utilize the semantic similarities between segments to transfer visual features. In the top row of Fig.~\ref{fig:attention}, attention in channels $1,4$ respectively perform transformations: $\texttt{sky\_other} \shortrightarrow \texttt{clouds}$, $\texttt{tree} \shortrightarrow \texttt{\{tree, hill\}}$. In the bottom row, attention in channels $1,2,7$ respectively perform transformations: $\texttt{clouds} \shortrightarrow \texttt{clouds}$, $\texttt{snow} \shortrightarrow \texttt{sand}$ and $\texttt{other} \shortrightarrow \texttt{\{surfboard,other\}}$.
% c) to transfer information to new semantic regions, spatial attention attends to the global image or combinations of multiple regions. For example, the transformation $\texttt{tree,snow} \shortrightarrow \texttt{mountain}$ in top row channel 5 and $\texttt{sky,snow} \shortrightarrow \texttt{sea}$ in bottom row channel 2.

% \noindent \textbf{The Effect of Masking} \quad
% \TODO{add figure}

\begin{figure}[t]
	\centering
	\includegraphics[width=0.95\linewidth]{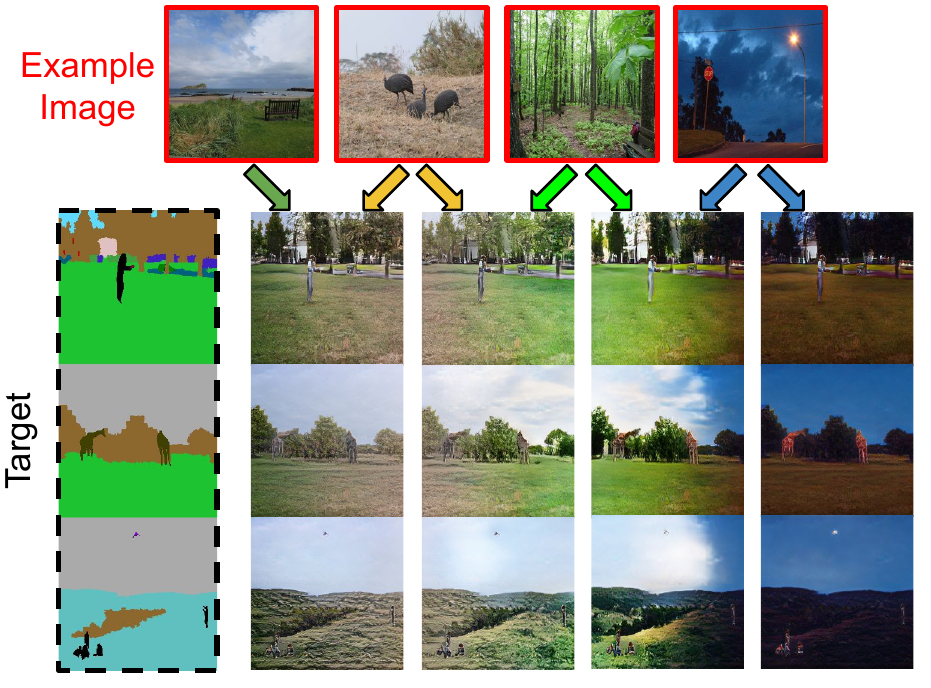}
	\caption{Spatially interpolate two styles in a single image at test stage. The styles of the synthesized images are deliberately changed from left to right.}
	\label{fig:interpolate_spatial}
\end{figure}

\noindent \textbf{Interpolation} \quad
We can easily control the synthesized styles in the test stage by manipulating attentions. Here, we show how to interpolate between two styles using our trained model: given two example images $x_2$ and $x_3$, we first compute their image features $F^{(i)}_{x,2},F^{(i)}_{x,3}$ and the spatial-attention maps $\alpha^{(i)}_2, \alpha^{(i)}_3$. Given an interpolating factor $\alpha\in[0,1]$ where $\alpha=1$ means ignoring the example scene $x_3$, the spatial attention map of the first scene is modified by $\alpha^{(i)}_2 \coloneqq \alpha^{(i)}_2+\log(\frac{\alpha^{(i)}_2}{1-\alpha^{(i)}_2})$. Afterwards, both feature maps $F^{(i)}_{x,2},F^{(i)}_{x,3}$ and spatial attention $\alpha^{(i)}_2, \alpha^{(i)}_3$ are concatenated along the horizontal axis. In addition, the masking score (output of the 2-layer MLP in Eq.~\ref{eq:masking}) is also interpolated. With the remaining procedures unchanged, i.e., same spatial aggregation, feature masking, channel aggregation and synthesis, interpolation results are readily generated. As shown in Fig.~\ref{fig:interpolate}, with slight modifications, our model can perform effective style interpolation. Specifically, the style traverses along the path $\texttt{grass} \shortrightarrow \texttt{dessert}\shortrightarrow \texttt{forest}\shortrightarrow \texttt{night}$ is achieved in Fig.~\ref{fig:interpolate}.

Likewise, by manipulating the channel attention at each spatial location, it is possible to adaptively mix style to synthesize an output image, i.e. spatial styles interpolation. As shown in Figure~\ref{fig:interpolate_spatial}, using the previous input, we interpolate between styles from left to right in a single image.

\noindent \textbf{Extrapolation} \quad 
\begin{figure}[t]
	\centering
	\includegraphics[width=0.95\linewidth]{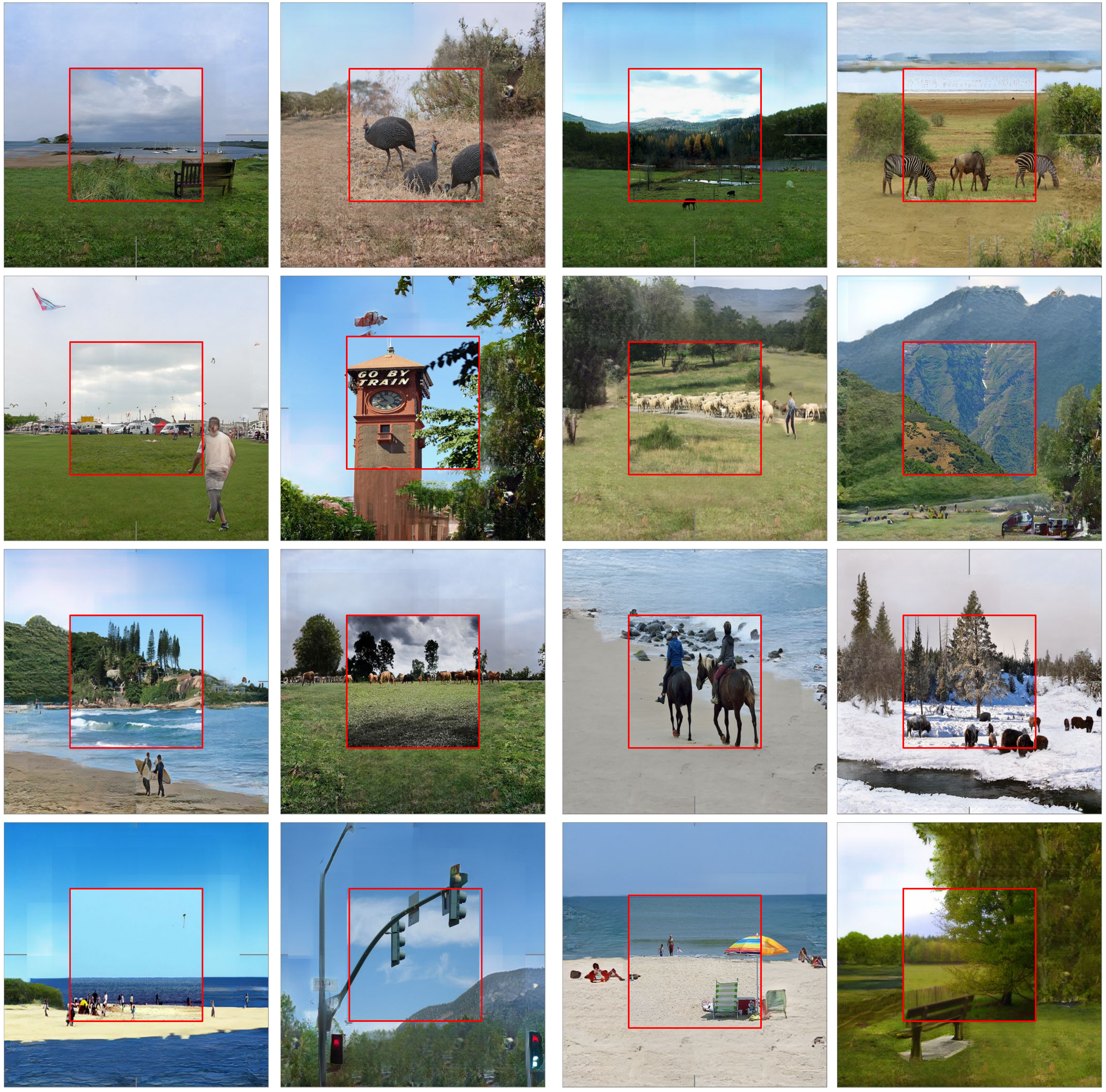}
	\caption{Given a scene patch at center, our model can generate
	Scene extrapolation based on patch.}
	\label{fig:extrapolation}
\end{figure}
Given a scene patch at the center, our model can achieve scene extrapolation, i.e.  generating beyond-the-border image content according to the semantic map guidance. A $512\times 512$ extrapolated images is generated by weighted combining synthesized $256\times256$ patches at $4$ corners and $10$ other random locations. As shown in Fig.~\ref{fig:extrapolation}, our model generates visually plausible extrapolated images, showing the promise of our proposed framework for guided scene panorama generation.

\begin{figure}[t]
	\centering
	\includegraphics[width=1.0\linewidth]{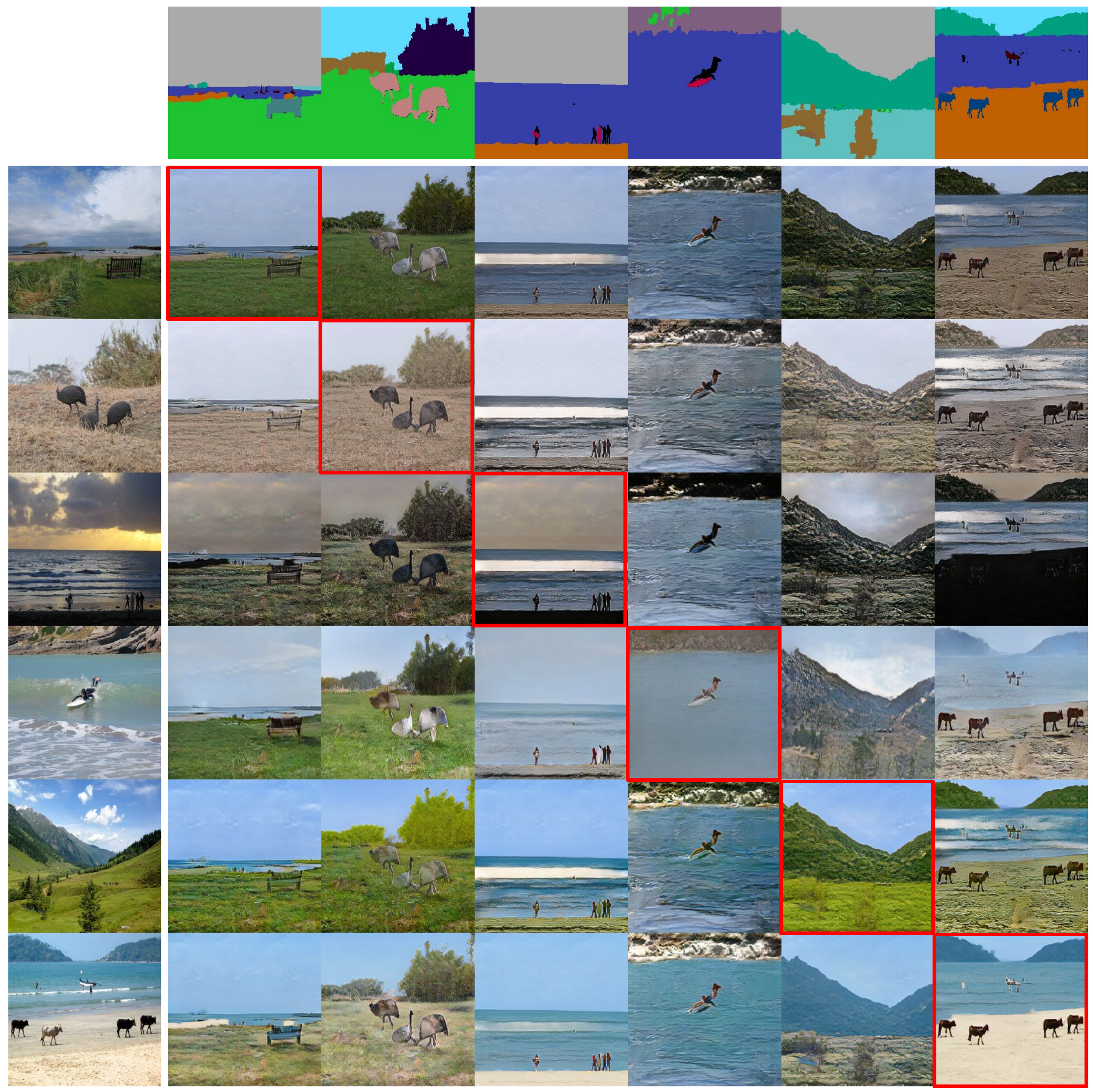}
	\caption{Style-structure swapping on 6 semantically unaligned arbitrary scenes at resolution $256\times 256$.
            % 	For each output in the $8\tiems 8$ grid, the corresponding segmentation map and style inputs are shown on top row and left column, respectively. 
            Our model can generalize across very different scene semantics and synthesize images with reasonable and consistent styles. Note that the images along the diagonal (red boxes) are {\it self-reconstruction}, which are generally quite good. Please zoom in for details.}
	\label{fig:swap1}
\end{figure}

\noindent \textbf{Swapping Style} \quad
Fig.~\ref{fig:swap1} shows reference-guided style swapping on six distinctively different scenes. For the same segmentation mask, we generate multiple outputs using different reference images. Our approach can reasonably transfer styles among multiple scenes, including grassland, dessert, ocean view, ice land, etc. More results are included in the appendix.

\section{Conclusion}
We propose to address a challenging example-guided scene image synthesis task. To propagate information between structurally uncorrelated and semantically unaligned scenes, we propose an MSCA module that leverages decoupled cross-attention for adaptive correspondence modeling. With MSCA, we propose a unified model for joint global-local alignment and image synthesis. We further propose a patch-based self-supervision scheme that enables training. Experiments on the COCO-stuff dataset show significant improvements over the existing methods. Furthermore, our approach provides interpretability and can be extended to other content manipulation tasks.

{\small
\bibliographystyle{ieee_fullname}
\bibliography{egbib}
}

\begin{appendices}
\section{The Synthesis Module}
\label{sec:synthesis}
\begin{figure*}[h]
	\centering
	\includegraphics[width=1.0\linewidth]{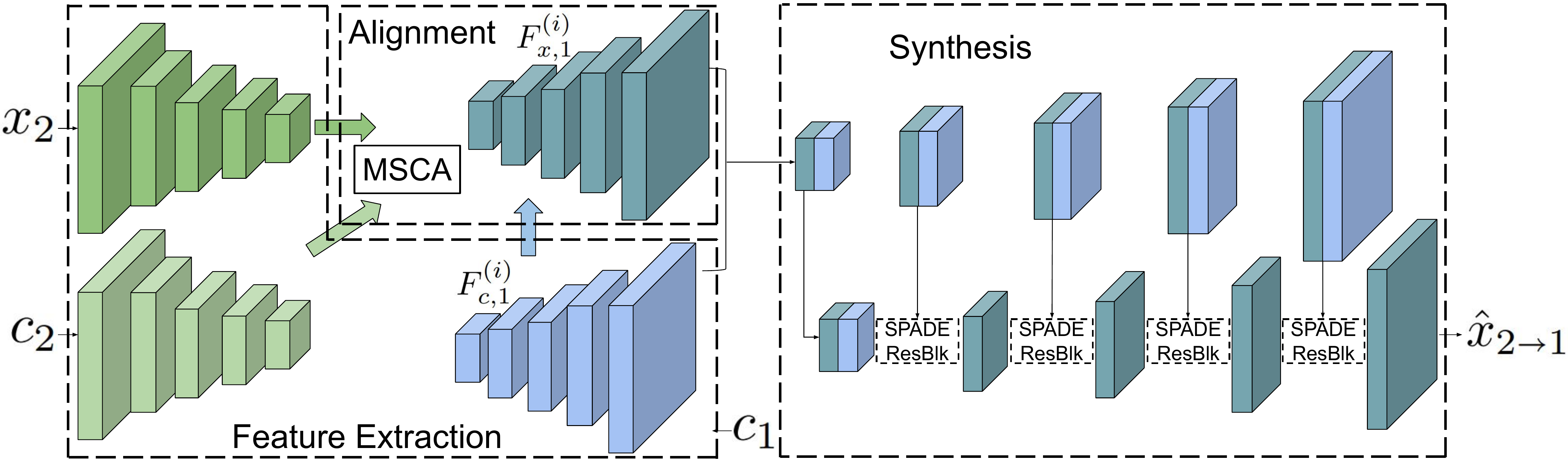}
	\caption{The details of the image synthesis module (the dash block on the right). The image synthesis module takes image features maps $F^{(i)}_{x,1}$ and segmentation features maps $F^{(i)}_{c,1}$ at all scale $i$ as inputs to output a new image $\hat{x}_{2\shortrightarrow 1}$. Multiple SPADE residue blocks~\cite{spade} with upsampling layers are used to upsample the spatial resolutions.
	% \todo{change figure notation, (image, segmap, all features's name, ), element-wise multiplication, add spatial channel attention map to figure to illstrate. }
	}
	\label{fig:sup-synthesis}
\end{figure*}

\noindent As shown in Fig.~\ref{fig:sup-synthesis}, our image synthesis module (the dash block on the right) takes the image features map $F^{(i)}_{x,1}$ and segmentation features map $F^{(i)}_{c,1}$ as inputs to output a new image $\hat{x}_{2\shortrightarrow 1}$. Specifically, at each scale, a SPADE residue block~\cite{spade} with upsampling layer takes the concatenation of $F^{(i)}_{x,1}$ and $F^{(i)}_{c,1}$ as input to generate an upsampled feature map or image. 

\section{MSCA Pretraining}
\label{sec:pretraining}
\begin{figure*}[h]
	\centering
	\includegraphics[width=0.9\linewidth]{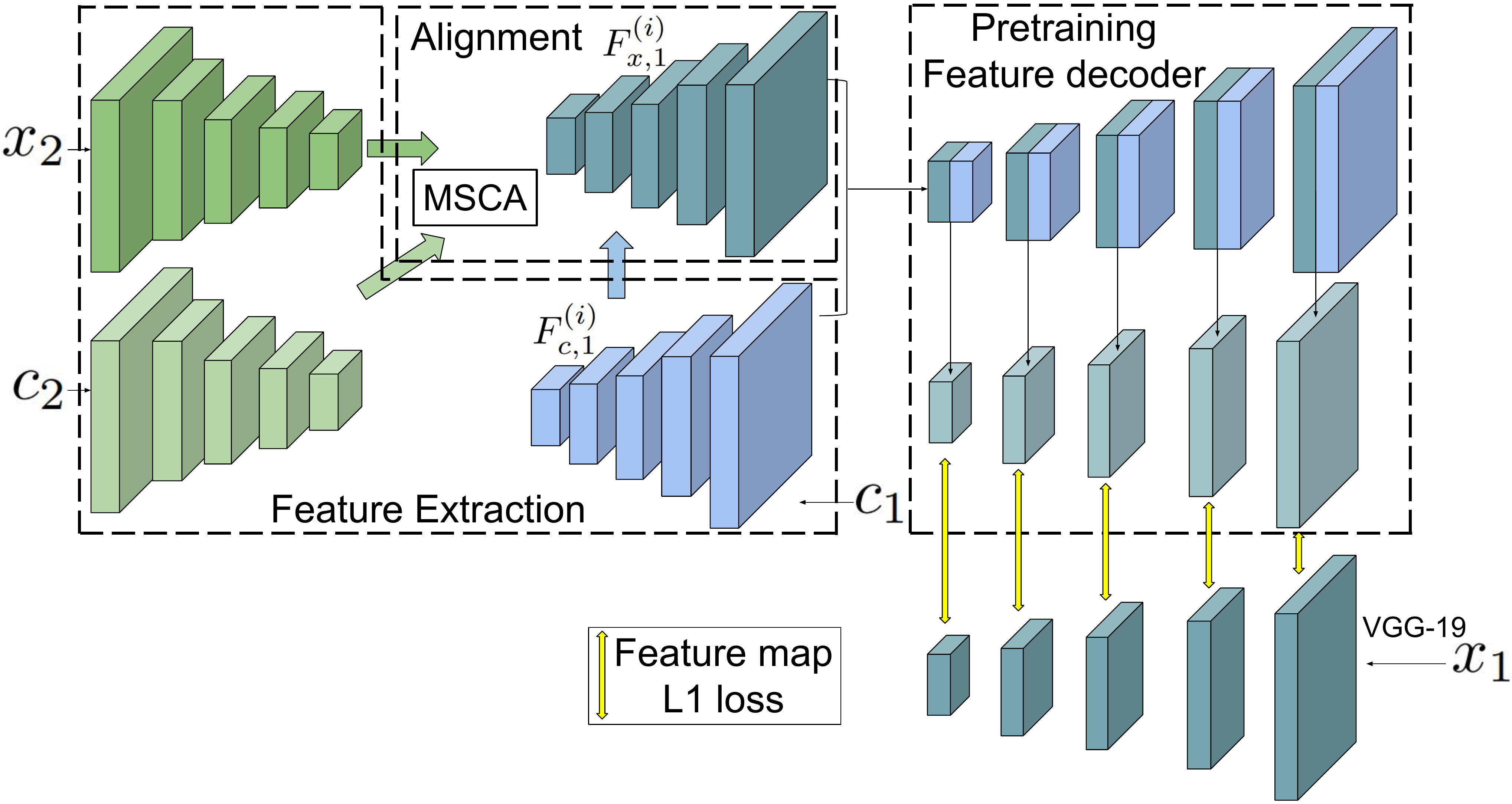}
	\caption{The details of the auxiliary feature decoder for feature extractor and MSCA pretraining (dash block on the right). At each scale $i$, the image features map $F^{(i)}_{x,1}$ and the segmentation features map $F^{(i)}_{c,1}$ are concatenated and feed to a $1\times 1$ convolution layer to predict the VGG-19 features map of $x_1$.
	}
	\label{fig:pretraining}
\end{figure*}

\noindent As shown in Fig.~\ref{fig:pretraining}, an auxiliary feature decoder (the dash block on the right) is used to pretrain the feature extractors and the MSCA modules. Specifically, at each scale, the concatenation of $F^{(i)}_{x,1}$ and $F^{(i)}_{c,1}$ at each scale is fed into a $1\times 1$ convolutional layer to reconstruct the ground-truth VGG feature of $x_1$ at the corresponding scale. We weighted sum the L1 losses between predictions and ground-truth at each scales, then apply backpropagation to update weights of the whole model. We pretrain the model for $20$ epochs. Because of the light-weight design of the feature decoder, the pretraining step only takes around $12$ hours, and around ~$4$\% of the total training time.

\section{Results of \cite{example_cvpr18,example_cvpr19}}
\label{sec:compare_cvpr18_19}
\begin{figure*}[t]
	\centering
	\includegraphics[width=0.9\linewidth]{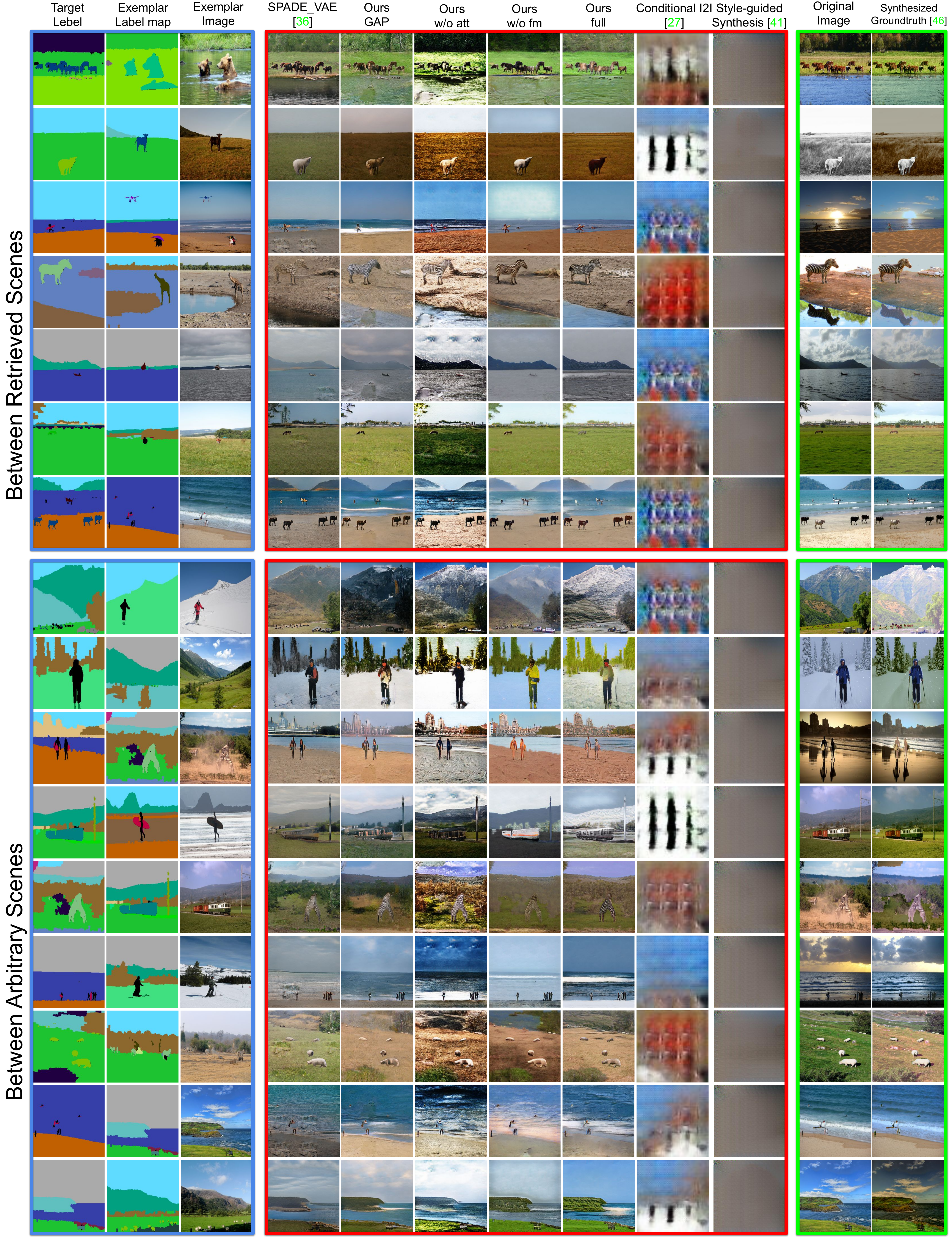}
	\caption{More visual comparisons to \cite{example_cvpr18} and \cite{example_cvpr19} (at columns 9 and 10, respectively).
	Example-guided scene synthesis is performed on {\bf{retrieved scenes}} (top) and {\bf{arbitrary scenes}} (buttom).  Columns 1 to 3 (blue) depict the target label maps, exemplar label maps and the associated images, respectively. Columns 4 to 8 in (red) depict different methods and our model (Columns 8).  
	Columns 13 and 14 (green) respectively depict original images from target label map and synthesized ground truth using~\cite{wct2} in the \emph{retrieved} dataset. Our method clearly produces the most style-consistent and visually plausible images.
% 	\todo{change blue box to other color, bolder box.}
	}
	\label{fig:compare_main_new}
\end{figure*}

\noindent We provide additional results of conditional image-to-image translation (Conditional I2I)~\cite{example_cvpr18} and style-guided synthesis~\cite{example_cvpr19} in Fig.~\ref{fig:compare_main_new}, column 9 and 10. To train the model of~\cite{example_cvpr18}, we resize images and semantic label maps to $64\time 64$, the original resolution used in~\cite{example_cvpr18}. We test different learning rates and early stopping strategies to prevent the generator from model collapse. To implement~\cite{example_cvpr19}, we train the model of ~\cite{example_cvpr19} using our patch-based self-supervision. We test multiple learning rates and channel sizes of the generator. However, we could not achieves good results for \cite{example_cvpr18} and \cite{example_cvpr19}. We believe the disentanglement strategy of \cite{example_cvpr18} is too challenging for the highly diversified COCO-stuff dataset. Meanwhile, input domain concatenation used in~\cite{example_cvpr19} may not be sufficient to capture and fuse the style information for the more challenging scene image dataset. In addition, spatially-adaptive normalization~\cite{spade} might be required for~\cite{example_cvpr19} to better utilize the captured style coding.

\vspace{-3px}
\section{More Style Swapping Results}
\label{sec:swap}
\begin{figure*}[]
	\centering
	\includegraphics[width=1.0\linewidth]{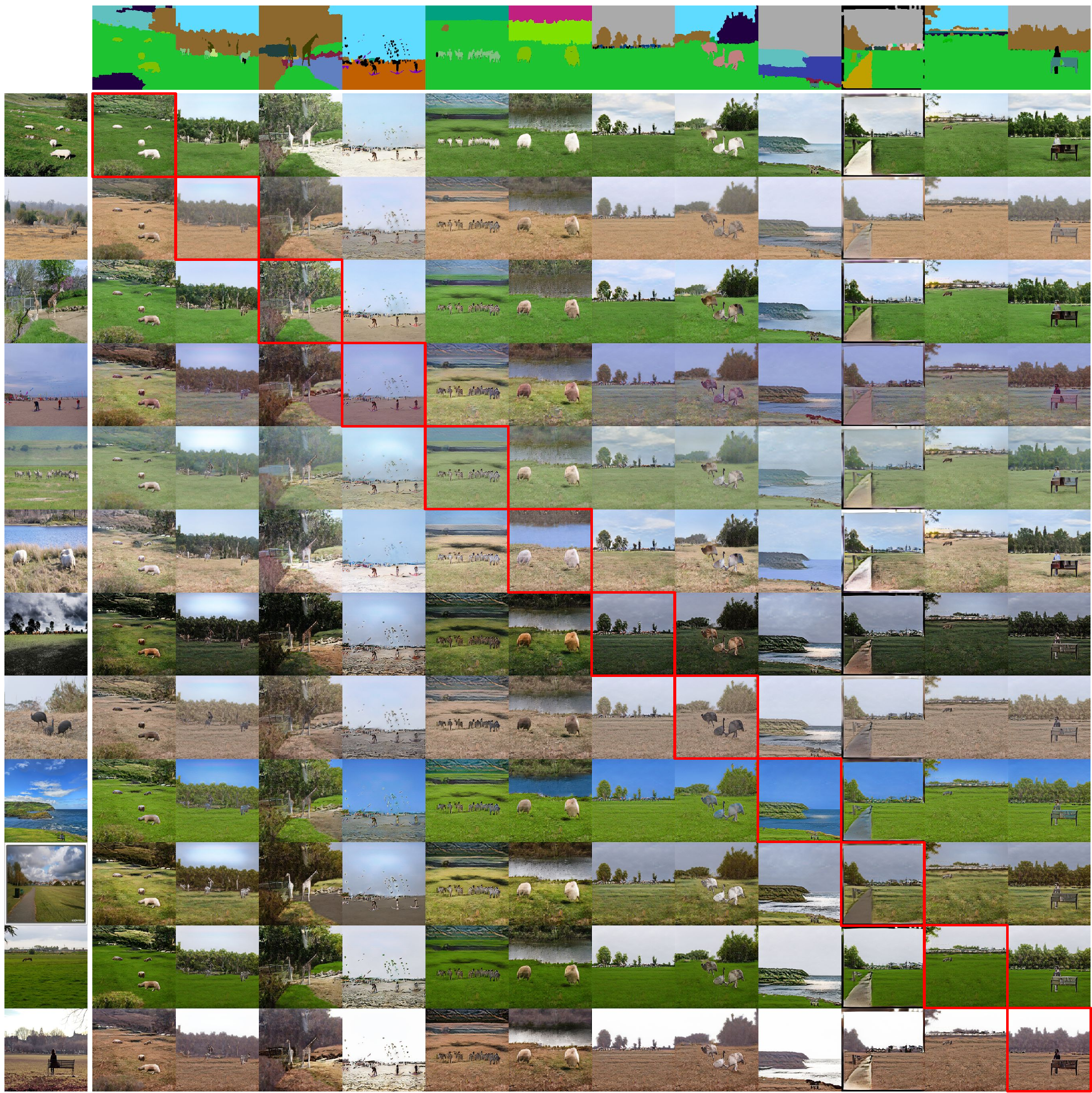}
	\caption{Style-structure swapping on 12 semantically unaligned and arbitrary scenes at resolution $256\times 256$.
            Our model can generalize across very different scene semantics and synthesize images with reasonable and consistent styles. Note that the images along the diagonal (red boxes) are {\it self-reconstruction}, which are generally quite good. Please zoom in for details.}
	\label{fig:swap12}
\end{figure*}

We show style swapping results on $12$ diversified scenes in Fig.~\ref{fig:swap12}. As shown in the figure, our model can transfer styles to very different scene semantics and generate style consistent outputs given exemplar images. 

\end{appendices}

\end{document}